\documentclass[conference]{IEEEtran}
\IEEEoverridecommandlockouts
\usepackage{cite}
\usepackage{amsmath,amssymb,amsfonts}
\usepackage{algorithmic}
\usepackage{graphicx}
\usepackage{textcomp}
\usepackage{xcolor}
\usepackage{float}
\usepackage{hyperref}
\hypersetup{
    colorlinks=true,
    linkcolor=cyan,
    filecolor=magenta,      
    urlcolor=blue,
}

\def\BibTeX{{\rm B\kern-.05em{\sc i\kern-.025em b}\kern-.08em
    T\kern-.1667em\lower.7ex\hbox{E}\kern-.125emX}}
\begin{document}

\title{Fast Interactive Video Object Segmentation with Graph Neural Networks

\thanks{\copyright 2021 IEEE.  Personal use of this material is permitted.  Permission from IEEE must be obtained for all other uses, in any current or future media, including reprinting/republishing this material for advertising or promotional purposes, creating new collective works, for resale or redistribution to servers or lists, or reuse of any copyrighted component of this work in other works.

Viktor Varga was supported by the Hungarian Government and co-financed by the European Social Fund (EFOP-3.6.3-VEKOP-16-2017-00002: EFOP-3.6.3-VEKOP-16-2017-00002, Integrated Program for Training New Generation of Scientists in the Fields of Computer Science. Andr\'as L\H{o}rincz was supported by the Thematic Excellence Programme (Project no. ED\_18-1-2019-0030 titled Application-specific highly reliable IT solutions) of the National Research, Development and Innovation Fund of Hungary. The authors thank Robert Bosch, Ltd. Budapest, Hungary for their generous support to the Department of Artificial Intelligence}
}

\author{\IEEEauthorblockN{Viktor Varga}
\IEEEauthorblockA{\textit{Faculty of Informatics} \\
\textit{E\"otv\"os Lor\'and University}\\
Budapest, Hungary \\
vv@inf.elte.hu}
\and
\IEEEauthorblockN{Andr\'as L\H{o}rincz}
\IEEEauthorblockA{\textit{Faculty of Informatics} \\
\textit{E\"otv\"os Lor\'and University}\\
Budapest, Hungary \\
lorincz@inf.elte.hu}
}

\maketitle

\begin{abstract}

Pixelwise annotation of image sequences can be very tedious for humans. Interactive video object segmentation aims to utilize automatic methods to speed up the process and reduce the workload of the annotators. Most contemporary approaches rely on deep convolutional networks to collect and process information from human annotations throughout the video. However, such networks contain millions of parameters and need huge amounts of labeled training data to avoid overfitting. Beyond that, label propagation is usually executed as a series of frame-by-frame inference steps, which is difficult to be parallelized and is thus time consuming.
In this paper we present a graph neural network based approach for tackling the problem of interactive video object segmentation. Our network operates on superpixel-graphs which allow us to reduce the dimensionality of the problem by several magnitudes. We show, that our network possessing only a few thousand parameters is able to achieve state-of-the-art performance, while inference remains fast and can be trained quickly with very little data.

\end{abstract}

\section{Introduction}

Video object segmentation (VOS) aims to segment multiple objects in videos from possibly unseen semantic categories, without identifying the category itself. The most popular ways to solve VOS are fully automatic (unsupervised) foreground segmentation and semi-supervised object segmentation methods \cite{vos_survey}. Both approaches are inherently limited and can be rarely applied in practice when high quality segmentation is needed. 

While unsupervised VOS solutions do not require human supervision, they easily fail in cases when the objects are not moving or are not clearly separated (e.g., a person riding a bicycle). Also, by definition \cite{davis_challenge17, youtube_vos}, it is not allowed for human annotators to indicate objects of interest in the scene, or to make corrections in the label predictions, therefore the maximum accuracy is limited.

\begin{figure*}
    \centering 
    \includegraphics[width=\textwidth]{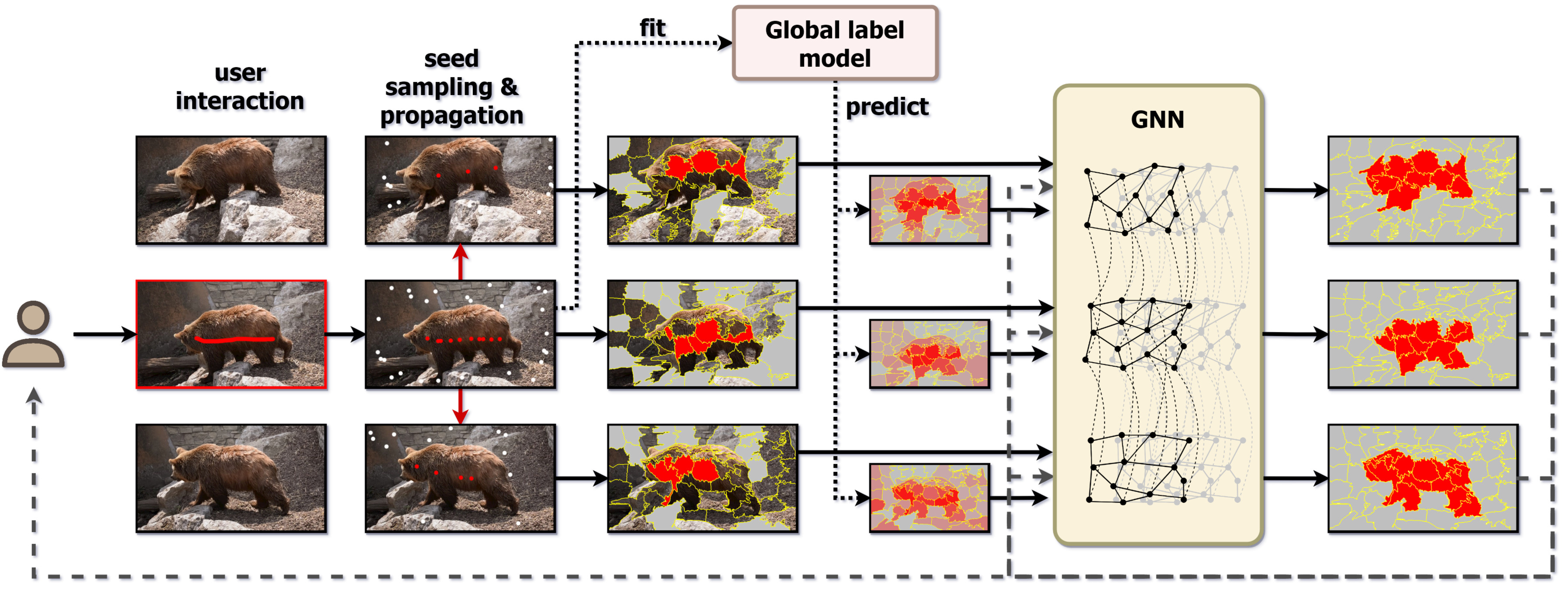}
    \caption{Overview of our method. First, the user submits a scribble annotation in one of the frames of the video. Seed points are sampled from the scribble. In the first interaction round, negative (background) seeds are generated as well. The global label model is trained from feature vectors extracted in these points. Then, the seed points are propagated into other frames of the video by our seed propagation algorithm. We annotate the nodes of the superpixel graph based on the location of the seed points. The propagated and the actual annotations form the input of the GNN together with the predictions of the label model, and the precomputed image and optical flow based features in all superpixel nodes. The GNN predicts the labeling of the superpixel nodes throughout the video. The prediction is then returned to the user for correction and it is also used as an input to the GNN in the next inference phase.}
    \label{fig:overview}
\end{figure*}

Semi-supervised VOS receives full, pixelwise GT annotation in one or several frames throughout the video. However, producing high quality annotation even for a single frame can be very slow \cite{davis_challenge18} and again, by definition \cite{davis_challenge17, youtube_vos} no corrections can be made in the label predictions of unannotated frames.

On the other hand, interactive VOS suits applications requiring 90+\% accuracy in segmentation labeling. In this case, precise high quality full frame annotations are replaced by much quicker, although less precise annotation means: the annotator iteratively marks mistakes in label predictions. Following each interaction step, the method updates the label predictions for the whole video, using the information provided by the annotator.

Acuna et al. \cite{interactive_polygonrnn} predict the outline of objects as polygons, which can be efficiently corrected by the dragging of their vertices. Song et al. \cite{interactive_seednet} aim to learn automatic seed point generation in image segmentation and also enable users to intervene by injecting seed points manually. Nagaraja et al. \cite{interactive_nagaraja_iccv15} employ scribbles - a crude, but fast way to input user annotation.

Caelles et al. \cite{davis_challenge18} put together the first and only database so far to support evaluation of interactive VOS methods with the use of scribbles and a simulated annotator agent. The benchmark automatically generates scribbles from the ground-truth label masks to correct submitted dense label predictions, in one frame at a time. In our paper, we use the same annotation technique to be able to derive quantitative comparisons between our works and other ones.

A large portion of recently published methods in interactive VOS exploit deep Convolutional Neural Networks (CNNs) and are implemented in two main steps. First, the scribble input of each interaction step and predictions obtained from the previous interaction steps are used to estimate the mask and the pixelwise labeling. Second, the estimated mask is propagated in both directions along the video and the label predictions for each frame are updated. This approach has several disadvantages. The mask propagation and the repeated estimation in consecutive frames are computationally intensive and such tasks are restricted to sequential execution since propagation must move frame-by-frame through the video, preventing parallelization and giving rise to slow execution and long waiting times for the user. In addition, information propagation is unidirectional within a single interaction step. Finally, both mask estimation and mask propagation utilize deep CNNs having millions of parameters and enormous amounts of labeled data may be required for the proper training of these networks to avoid overfitting. Further review of the literature follows later.

By contrast, our method is based on a superpixel segmentation, from which a graph is constructed. Labeling of the graph is executed by a Graph Neural Network (GNN). The dimensionality reduction provided by the superpixel segmentation allows us to build models with several magnitudes fewer parameters (in the order of a few thousands only). Accordingly, our model is much less sensitive to overfitting and can be trained with a few videos of the DAVIS training set, while it still achieves high accuracy. Furthermore, the network operates on the whole image sequence at once and executes an iteration of information passing between nodes in each layer, enabling bidirectional information propagation.

In this paper we present the first GNN based solution to interactive VOS to our best knowledge, and reach state-of-the-art results. Our method can be trained with small amounts of data, without overfitting. Additionally, our method is fast, both in terms of training and inference. We release the software code and the trained model at \url{https://github.com/vvarga90/gnn_annot}.

\section{Related Work}

\subsection{Graph Neural Networks}

The extension of neural networks to process graph-structured data was first conceived by Gori et al. \cite{gnn_2005}. Scarselli et al. improved the method in 2009 \cite{gnn_2009}. Since then, many architectural variants have been proposed \cite{gnn_survey18, gnn_survey20} and customized for graph-, edge-, or node classification and regression \cite{gnn_graphsage, gnn_gat, gnn_gated_gcn}. In recent years, graph neural networks (GNNs) became highly successful in computer vision applications, too. Gao et al. \cite{gnn_gao_tracking19} integrate GNNs into a Siamese framework to solve single object tracking. Bras\'o and Leal-Taix\'e \cite{gnn_braso_tracking} utilize GNNs to find target correspondences in multi-object tracking. Wang et al. \cite{gnn_wang_actionrec18} proposes a somewhat similar approach for action recognition. Sarlin et al. \cite{gnn_sarlin_superglue} build their GNNs from alternating layers of self-attention and cross-attention and significantly improve accuracy and speed of feature matching in a pair of images.

\subsection{Video Object Segmetation}

\newcommand\figqualwidthratioM{0.19}
\newcommand\figqualgthspaceM{0.1}

\begin{figure*}[!ht]
\centering
\setlength{\tabcolsep}{1pt}
\begin{tabular}{ccccc}
GT & 1st step & 2nd step & 4th step & 8th step

\vspace{0.2cm}
\\

\includegraphics[width = \figqualwidthratioM\linewidth]{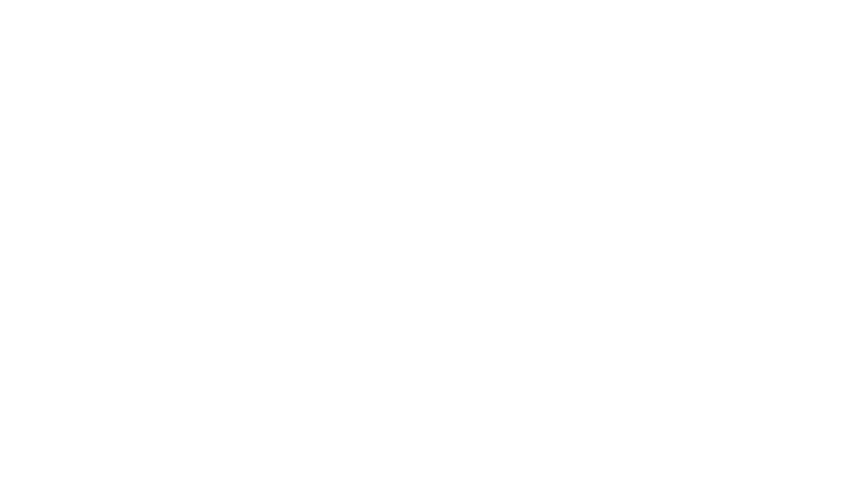} & 
\hspace{\figqualgthspaceM cm}
\includegraphics[width = \figqualwidthratioM\linewidth]{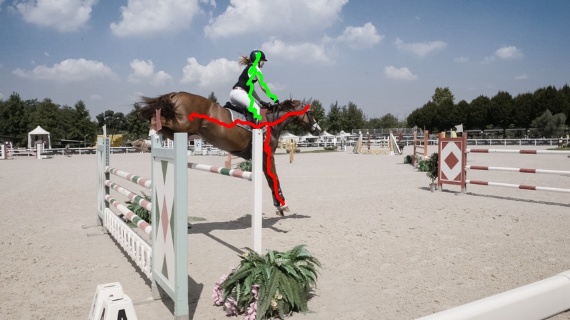} & 
\includegraphics[width = \figqualwidthratioM\linewidth]{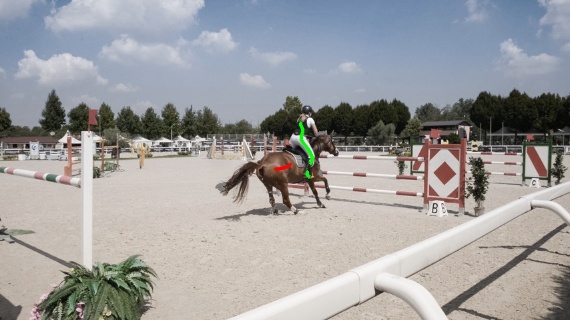} & 
\includegraphics[width = \figqualwidthratioM\linewidth]{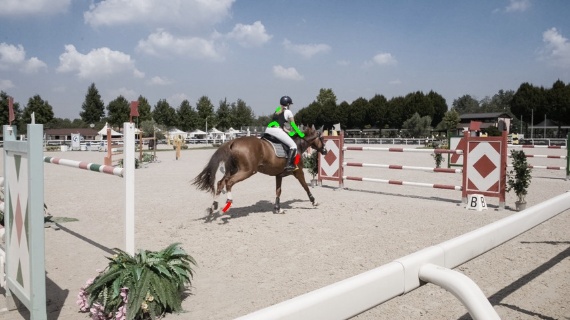} & 
\includegraphics[width = \figqualwidthratioM\linewidth]{img/qual_main/horsejump_high/horsejumphigh_scrib1_8_00000.jpg}\\

\includegraphics[width = \figqualwidthratioM\linewidth]{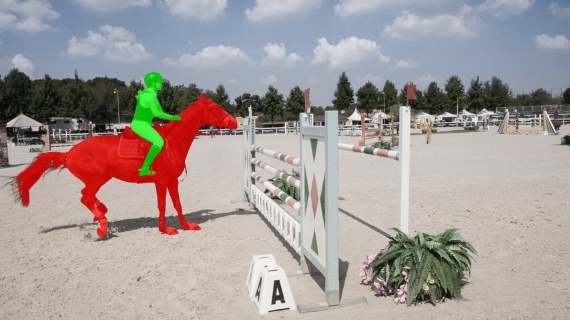} & 
\hspace{\figqualgthspaceM cm}
\includegraphics[width = \figqualwidthratioM\linewidth]{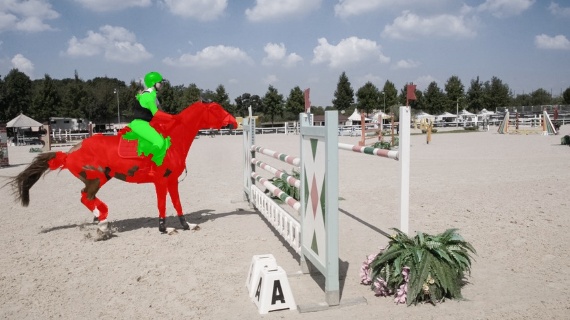} & 
\includegraphics[width = \figqualwidthratioM\linewidth]{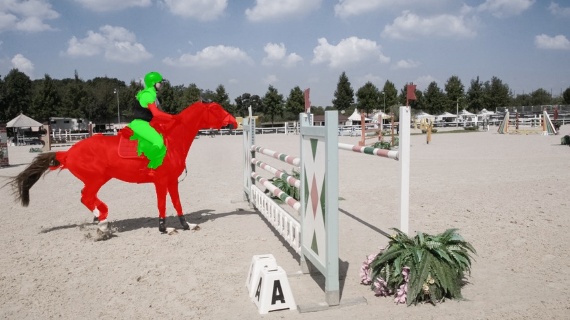} & 
\includegraphics[width = \figqualwidthratioM\linewidth]{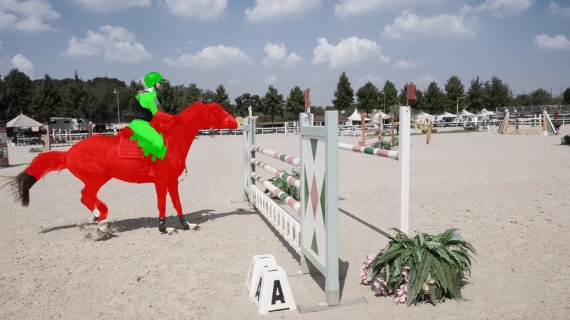} & 
\includegraphics[width = \figqualwidthratioM\linewidth]{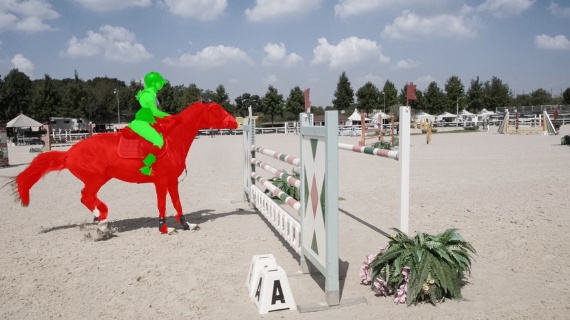}\\

\includegraphics[width = \figqualwidthratioM\linewidth]{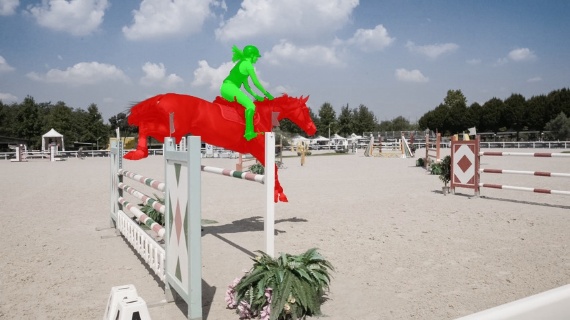} & 
\hspace{\figqualgthspaceM cm}
\includegraphics[width = \figqualwidthratioM\linewidth]{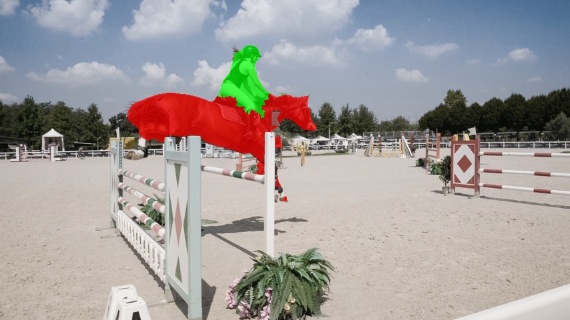} & 
\includegraphics[width = \figqualwidthratioM\linewidth]{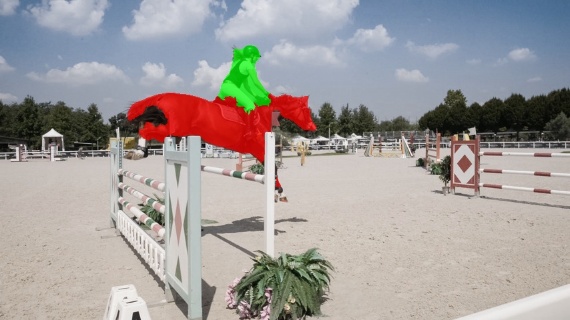} & 
\includegraphics[width = \figqualwidthratioM\linewidth]{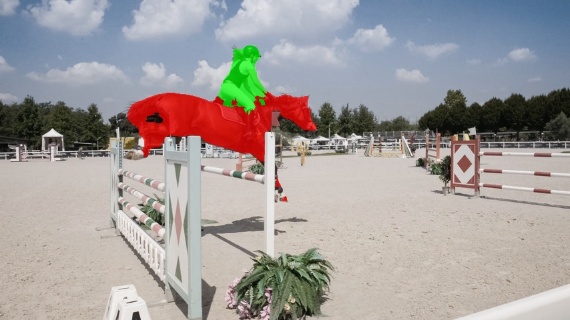} & 
\includegraphics[width = \figqualwidthratioM\linewidth]{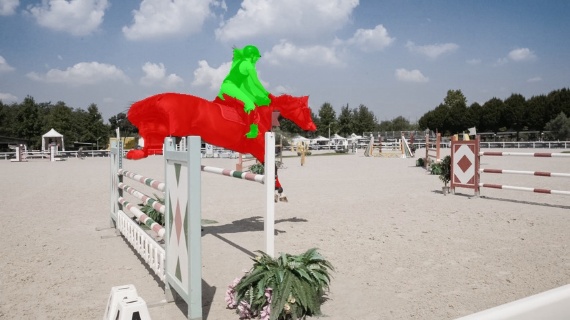}\\

\includegraphics[width = \figqualwidthratioM\linewidth]{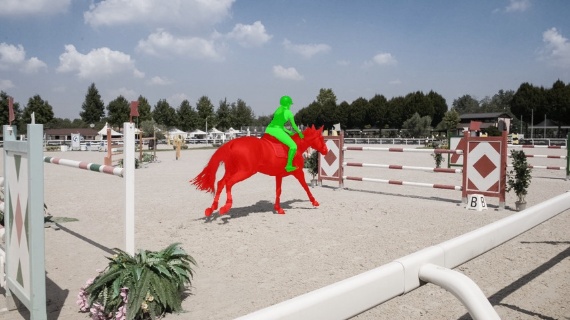} & 
\hspace{\figqualgthspaceM cm}
\includegraphics[width = \figqualwidthratioM\linewidth]{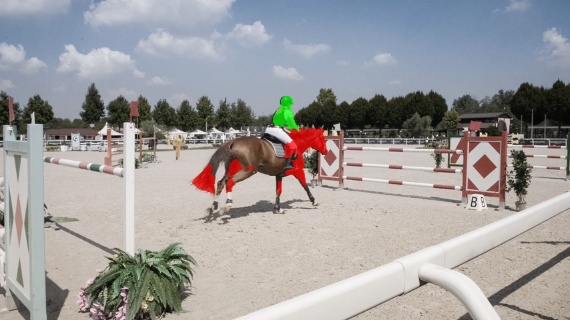} & 
\includegraphics[width = \figqualwidthratioM\linewidth]{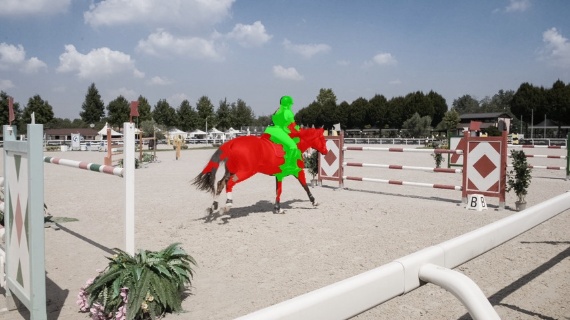} & 
\includegraphics[width = \figqualwidthratioM\linewidth]{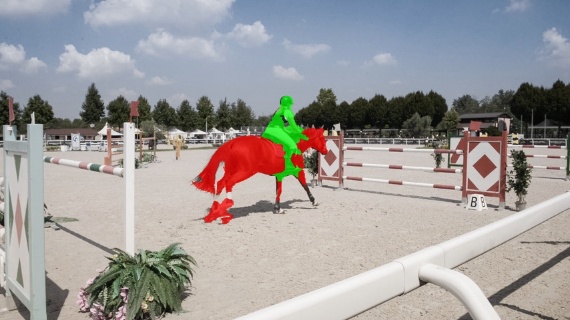} & 
\includegraphics[width = \figqualwidthratioM\linewidth]{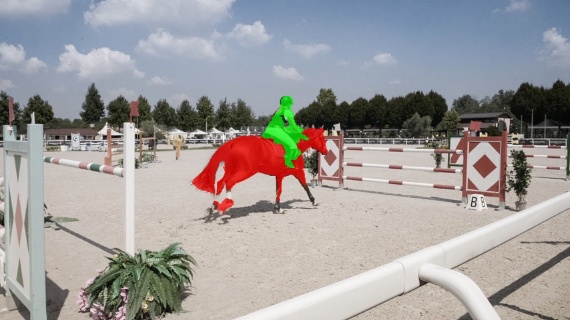}

\\
\\

\includegraphics[width = \figqualwidthratioM\linewidth]{img/qual_main/white.png} & 
\hspace{\figqualgthspaceM cm}
\includegraphics[width = \figqualwidthratioM\linewidth]{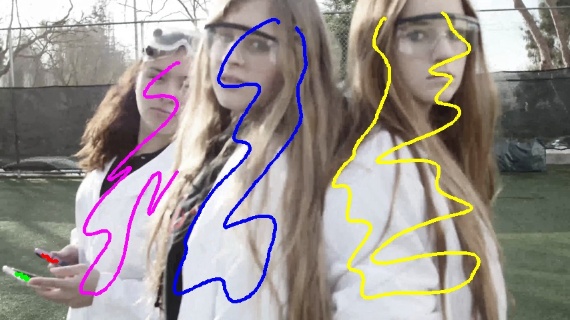} & 
\includegraphics[width = \figqualwidthratioM\linewidth]{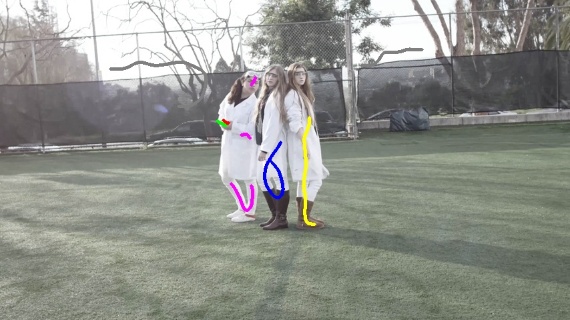} & 
\includegraphics[width = \figqualwidthratioM\linewidth]{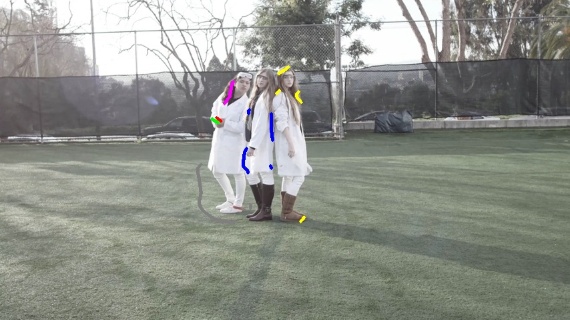} & 
\includegraphics[width = \figqualwidthratioM\linewidth]{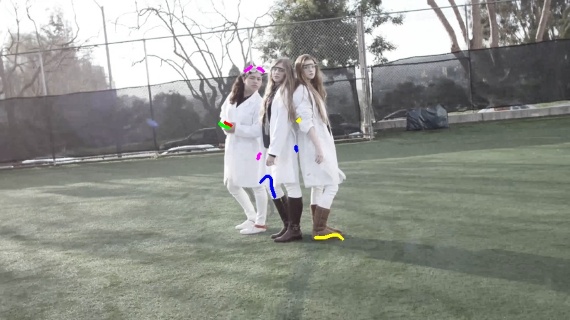}\\

\includegraphics[width = \figqualwidthratioM\linewidth]{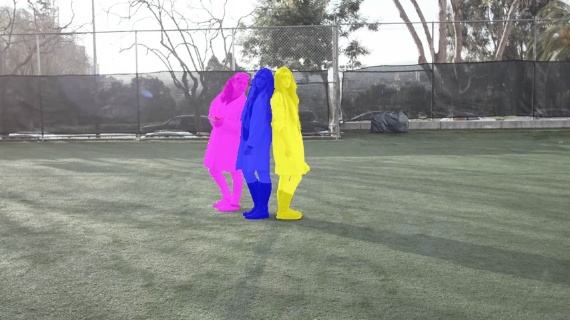} & 
\hspace{\figqualgthspaceM cm}
\includegraphics[width = \figqualwidthratioM\linewidth]{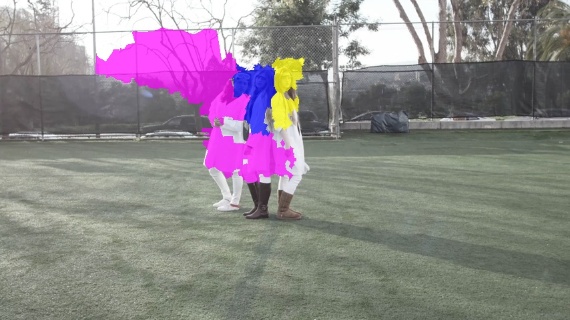} & 
\includegraphics[width = \figqualwidthratioM\linewidth]{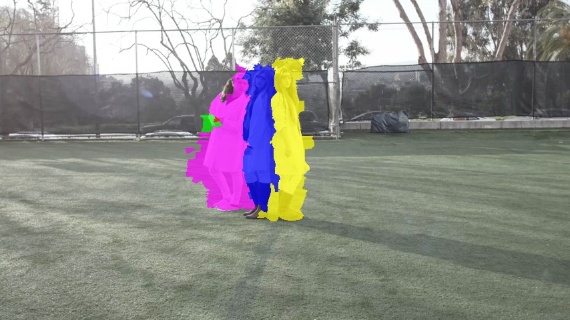} & 
\includegraphics[width = \figqualwidthratioM\linewidth]{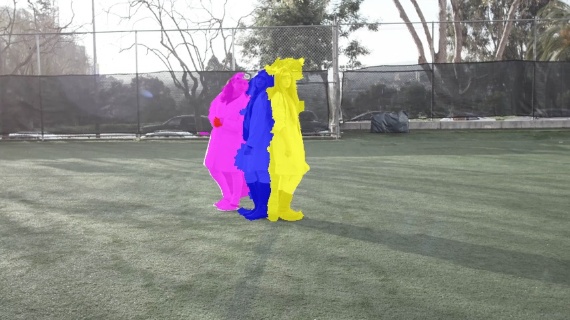} & 
\includegraphics[width = \figqualwidthratioM\linewidth]{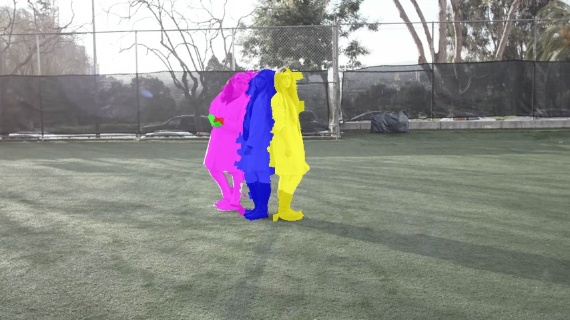}\\

\includegraphics[width = \figqualwidthratioM\linewidth]{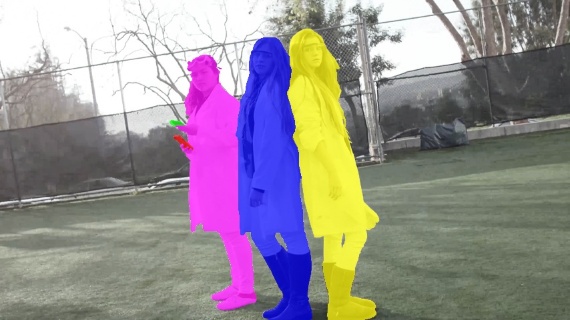} & 
\hspace{\figqualgthspaceM cm}
\includegraphics[width = \figqualwidthratioM\linewidth]{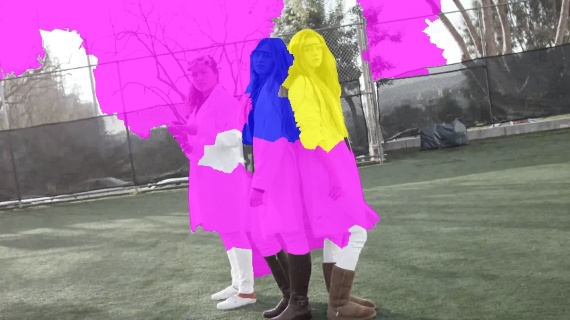} & 
\includegraphics[width = \figqualwidthratioM\linewidth]{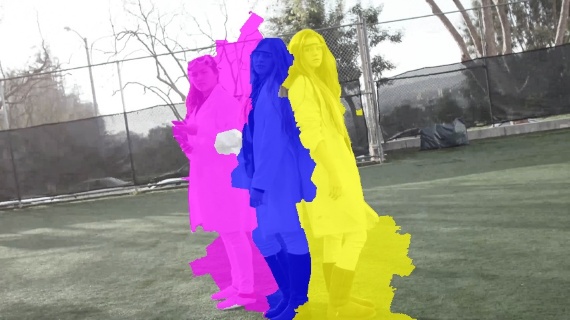} & 
\includegraphics[width = \figqualwidthratioM\linewidth]{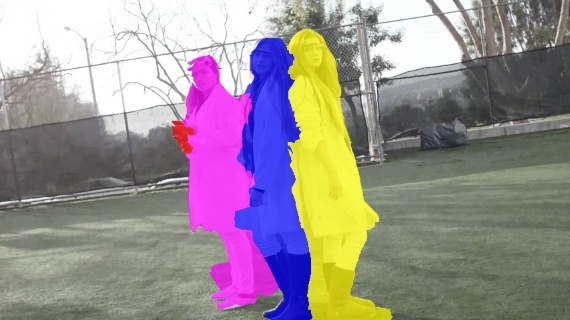} & 
\includegraphics[width = \figqualwidthratioM\linewidth]{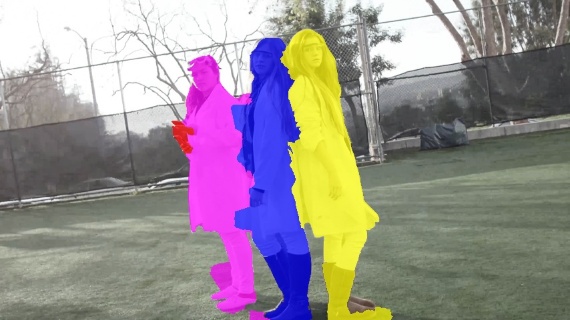}\\

\includegraphics[width = \figqualwidthratioM\linewidth]{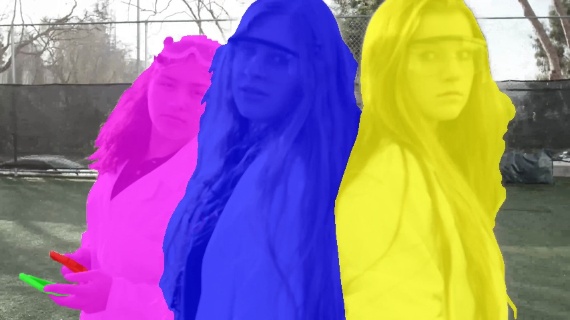} & 
\hspace{\figqualgthspaceM cm}
\includegraphics[width = \figqualwidthratioM\linewidth]{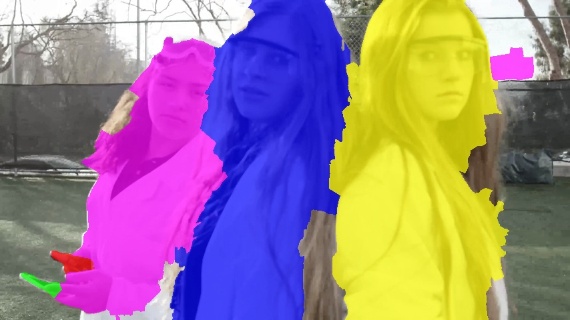} & 
\includegraphics[width = \figqualwidthratioM\linewidth]{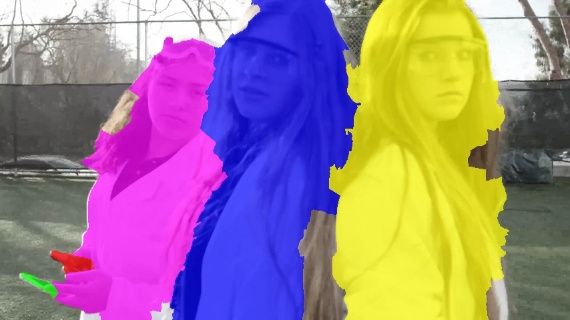} & 
\includegraphics[width = \figqualwidthratioM\linewidth]{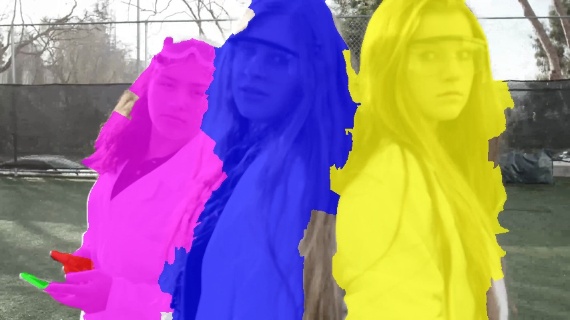} & 
\includegraphics[width = \figqualwidthratioM\linewidth]{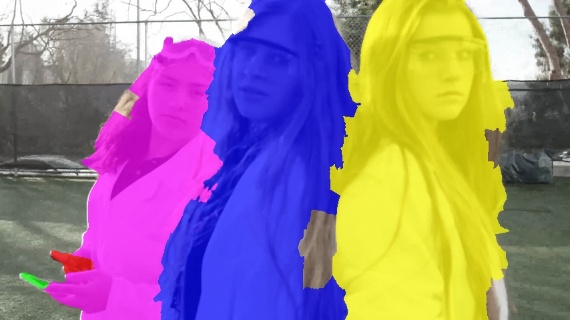}\\

\end{tabular}
\caption{Qualitative results with error rates from the \emph{worst quartile} of the DAVIS 2017 validation set. The top four rows and the bottom four rows show results on the \textit{horsejump-high} and the \textit{lab-coat} sequences, respectively. The first row for each video depicts the scribbles given by the simulated user in the corresponding interaction round. The 2nd to 4th  rows show label predictions in color for frames 0, 20, and 40 in this order. The 1st column contains the ground truth masks of the respective rows, whereas the 2nd to 5th columns show the predicted masks after the 1st, 2nd, 4th, and 8th iterations of user annotation inputs. Note the decrease of scribble sizes as time progresses due to the decrease of the erroneous regions. Best viewed in color.}
\label{fig:qual1}
\end{figure*}

Before the dominance of deep learning techniques, approaches to video object segmentation often included the handling of graphically structured data. Many solutions \cite{vos_classic_supervoxel1, vos_classic_supervoxel2, vos_classic_convex1} utilize superpixel (or supervoxel) segmentation algorithms as a preprocessing step to reduce the dimensionality of the image data. Superpixels \cite{superpixel_survey} provide a convenient way to identify spatio-temporally coherent regions of images and to reduce the dimensionality of the VOS problem.

In the past few years, end-to-end trained deep CNNs significantly raised the bar in the quality of segmentation. Introduction of larger scale and high resolution datasets, e.g., DAVIS \cite{davis_challenge17} and Youtube-VOS \cite{youtube_vos} further helped research. Both datasets released benchmarks in unsupervised and semi-supervised categories. In semi-supervised VOS, true pixelwise annotation is given in specific frames (usually in the first frame only) to define semantic categories of importance. Most solutions utilize CNNs in their pipelines. Khoreva et al. \cite{vos_semisup_lucid} present a refined data augmentation procedure to completely avoid pretraining of their models. Oh et al. propose the attention based Space-Time Memory networks \cite{vos_semisup_stm} which maintain a memory module to store feature information across time. Seong et al. refine \cite{vos_semisup_kernel2020} this method by prioritizing temporally local matches over more distant ones.

In unsupervised VOS, no indication is given about the semantic categories, resulting in many of the developed methods to seriously overfit to the dataset. Many solutions heavily rely on optical flow estimation to differentiate foreground objects from the background. Wang et al. utilize superpixel segmentation and graph-based saliency estimation \cite{vos_unsup_saliency} to solve unsupervised VOS. Goel et al. integrate the training of their video segmentation networks with that of a reinforcement learning agent to focus on segmentation of relevant objects \cite{vos_unsup_rl}. Oh et al. adapt \cite{vos_unsup_stm} Space-Time Memory networks to unsupervised VOS as well.

Graph-based solutions had been intermittently abandoned in the state-of-the-art literature, as GNN research lagged behind due to the lack of efficient software implementations. In the past two years, however, the field observed a dramatic growth that can be attributed partly to the introduction of graphically structured neural networks in popular deep-learning libraries \cite{pytorch, gnn_dgl}. Relatively few results combine GNNs with VOS as of yet. Wang et al. \cite{gnn_vos_zeroshot} propose a GNN variant and successfully matches object proposals in different frames of a video setting the new state-of-the-art in unsupervised VOS. Johnander et al. \cite{gnn_vos_unpub} applies a combination of recurrent and graph neural networks to perform real-time video instance segmentation. 

\subsection{Interactive Video Object Segmentation} \label{subsection:related_ivos}

Interactive image segmentation has a long history. Here, the human annotator and the machine take turns to refine the dense labeling of the image in cooperation. Well known algorithms, like GrabCut \cite{grabcut} continue to be in use as part of many annotation pipelines. A natural extension of the problem is interactive VOS, where the temporal propagation of information can drastically increase annotation efficiency.

Most interactive VOS methods follow a turn based annotation protocol: the task of annotation or correction of the predicted labeling in a certain frame or region is given. This task can be selected by either the user or the algorithm. Next, the user solves the task and these outputs are used in the improvement of the predicted segmentation labeling. Some methods \cite{activelearn_eccv12, activelearn_cvpr14, rlseg} intend to learn to select optimal annotation tasks for the user (sometimes called \textit{active learning}). A majority of approaches, including ours, focus on the efficient extraction of information from user input and try to maximize the improvement of the label predictions. Notable examples from the pre-deep learning era include \cite{interactive_livecut_iccv09, interactive_nagaraja_iccv15}.

Recently, deep learning based tools, such as deep CNNs have successfully been utilized for the task of interactive VOS as well. Quantitative analysis of published methods either relied on human experiments or used simulated annotators with unique annotation protocols. Both cases lead to difficulties in reproduction and comparison to other works. In 2018, Caelles et al. \cite{davis_challenge18} added an automatic interactive benchmark to the DAVIS video segmentation dataset. The benchmark follows the turn based scheme and for task selection the default evaluation algorithm is: the frame with the poorest quality labeling is selected for annotation. User input takes the form of one or several scribbles restricted to the frame being annotated.

Almost all top contenders of the DAVIS Interactive Challenges use deep CNNs. Training deep CNNs for mask estimation or propagation has multiple disadvantages. First, the number of trainable parameters is usually in the order of tens of millions, leading to serious overfitting for pixelwise labeling when only small amounts of data are available. Second, if deep CNNs propagate masks between adjacent frames then waiting time between interaction steps may be large since label prediction for each consecutive frame requires a full feed-forward pass of the network. Oh et al. solve \cite{interactive_oh_cvpr19} inveractive VOS with the help of two networks: an interaction network estimates segmentation masks from user scribbles, while a propagation network approximates labeling in consecutive frames. A slight deficiency of their method is the lack of weight sharing between the two networks in the convolutional layers. Heo et al. achieve superior results with their feature information transfer modules \cite{interactive20_1st}. A drawback of their method is the need to use multiple additional segmentation datasets for their training process. Miao et al. proposes a more efficient solution by computing all feature representations during the preprocessing stage and only utilizing shallow networks during prediction \cite{interactive_ma_net}. They implement global and local memory modules in a somewhat similar fashion to \cite{interactive20_1st}.

\newcommand\figqualwidthratioB{0.19}

\begin{figure*}[!ht]
\centering
\setlength{\tabcolsep}{1pt}
\begin{tabular}{ccccc}

\includegraphics[width = \figqualwidthratioB\linewidth]{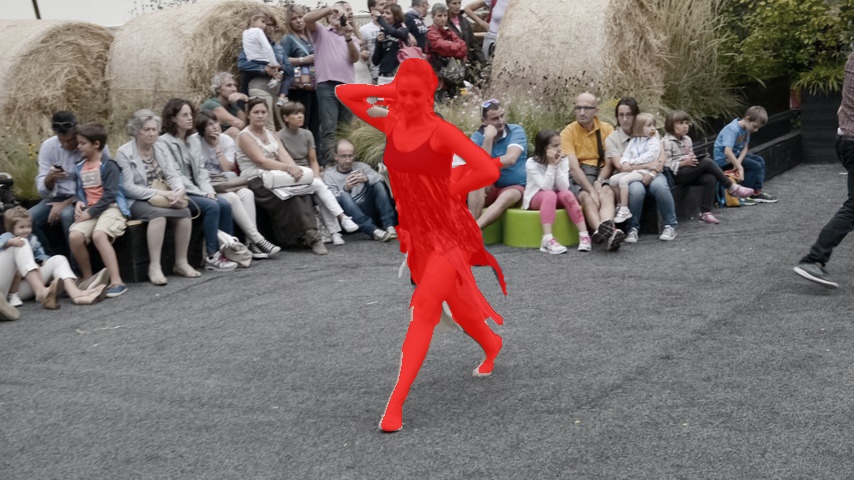} & 
\includegraphics[width = \figqualwidthratioB\linewidth]{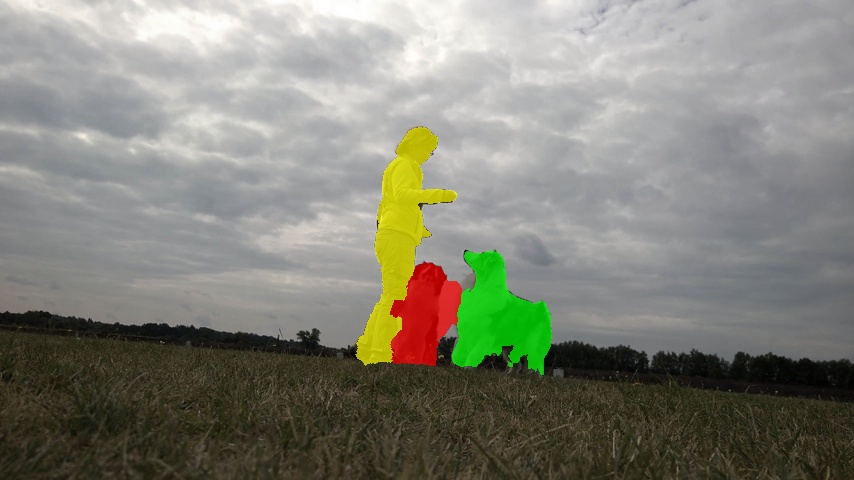} & 
\includegraphics[width = \figqualwidthratioB\linewidth]{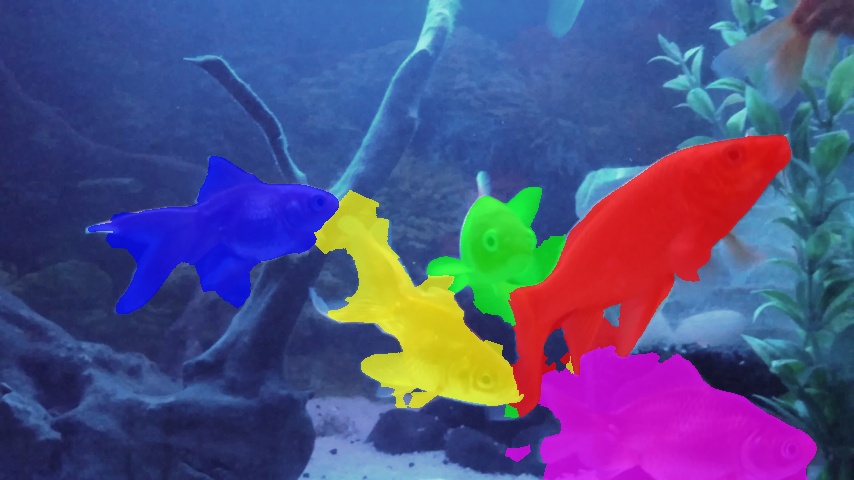} & 
\includegraphics[width = \figqualwidthratioB\linewidth]{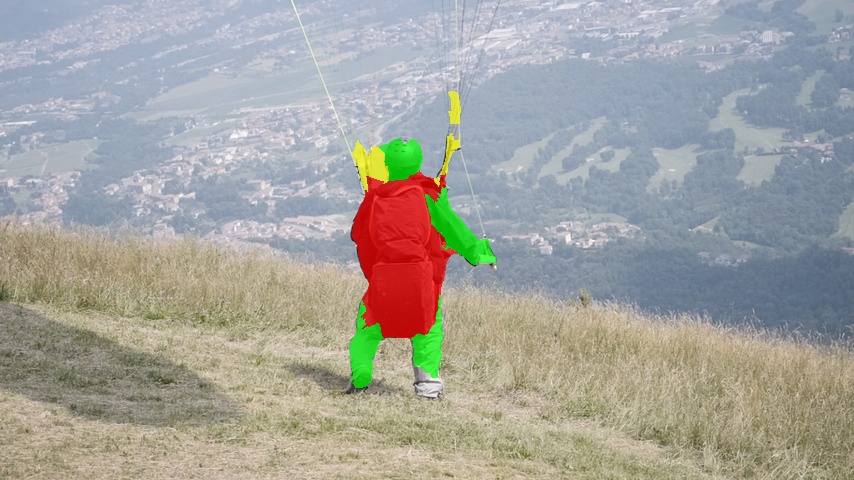} & 
\includegraphics[width = \figqualwidthratioB\linewidth]{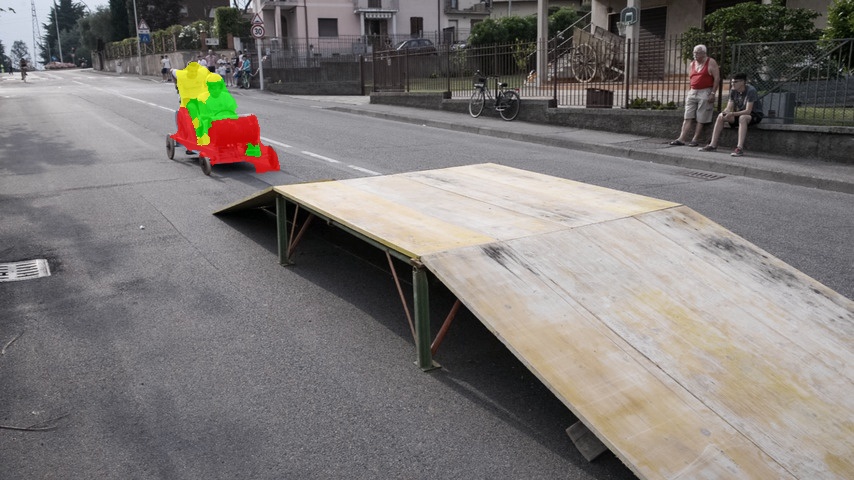}\\

\includegraphics[width = \figqualwidthratioB\linewidth]{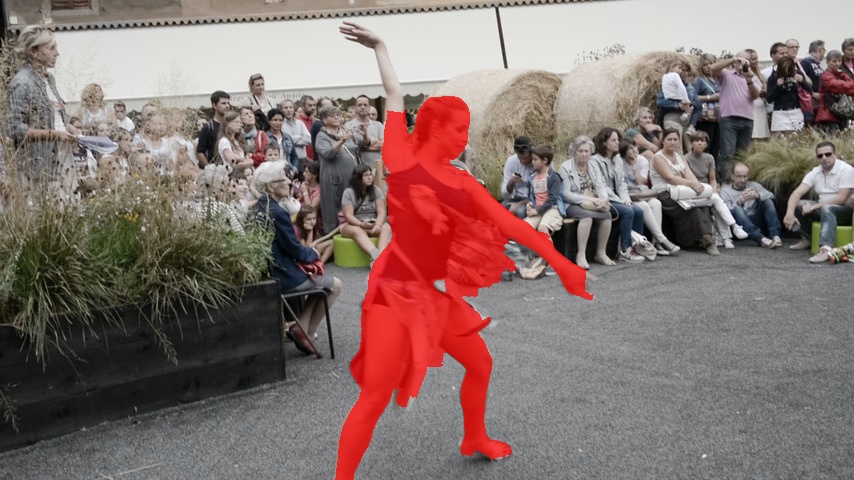} & 
\includegraphics[width = \figqualwidthratioB\linewidth]{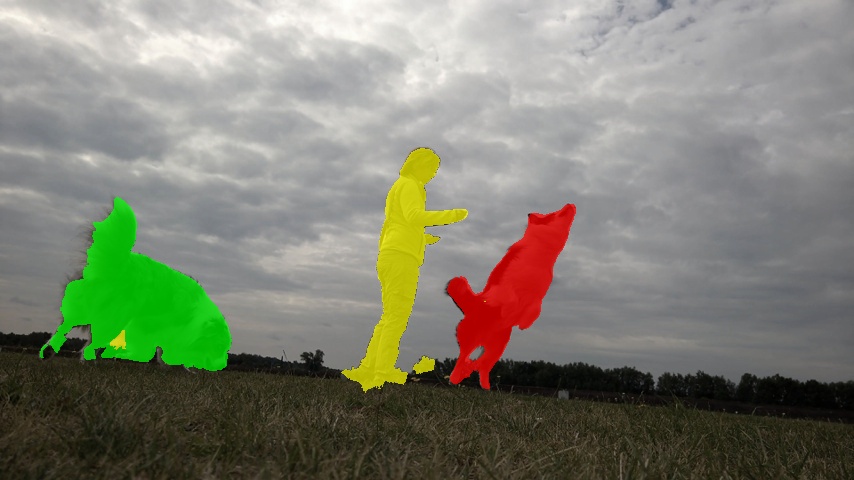} & 
\includegraphics[width = \figqualwidthratioB\linewidth]{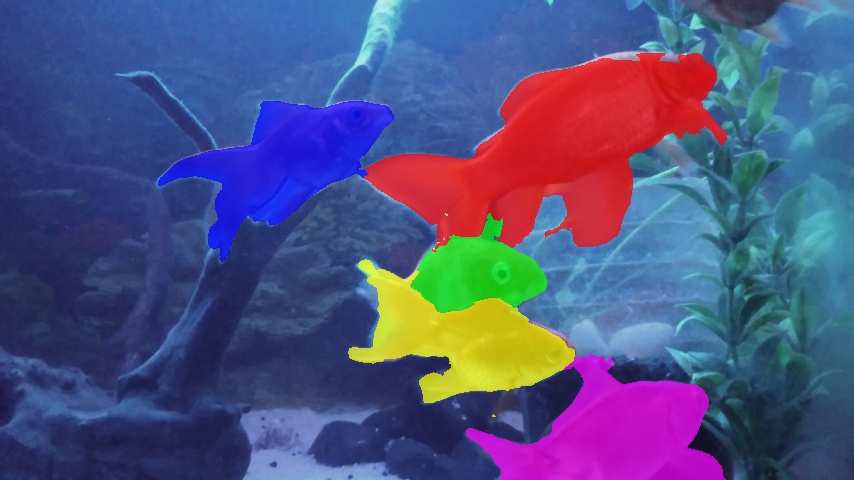} & 
\includegraphics[width = \figqualwidthratioB\linewidth]{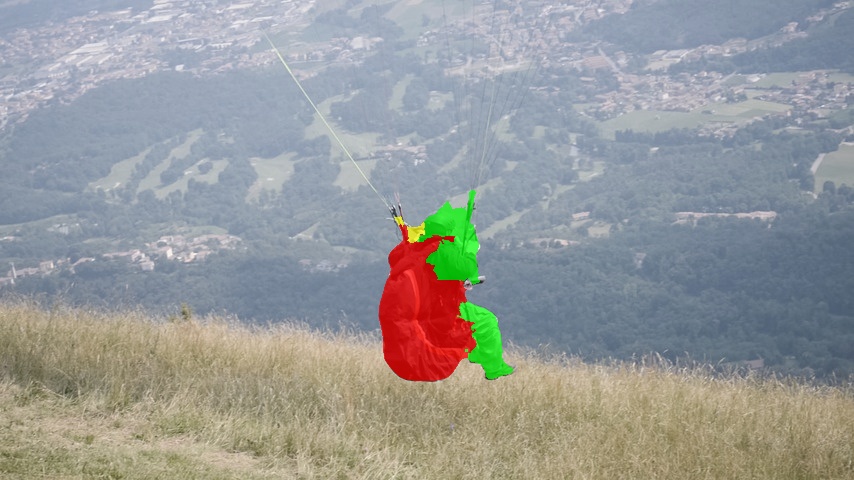} & 
\includegraphics[width = \figqualwidthratioB\linewidth]{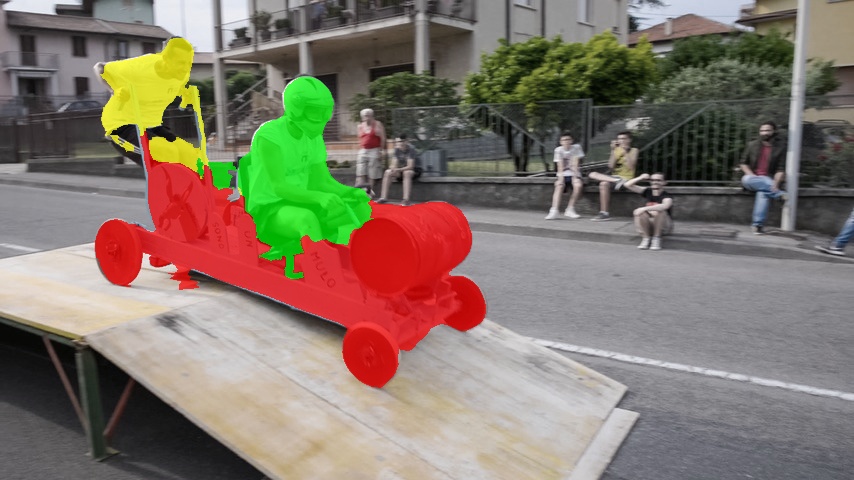}\\

\includegraphics[width = \figqualwidthratioB\linewidth]{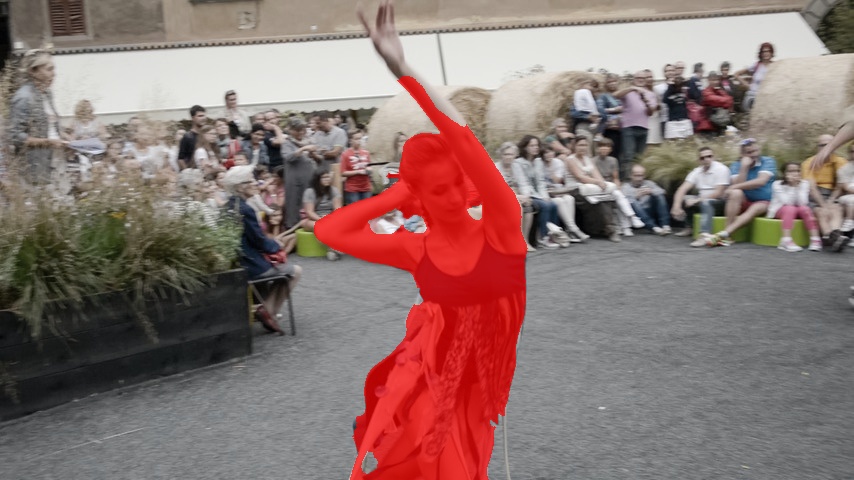} & 
\includegraphics[width = \figqualwidthratioB\linewidth]{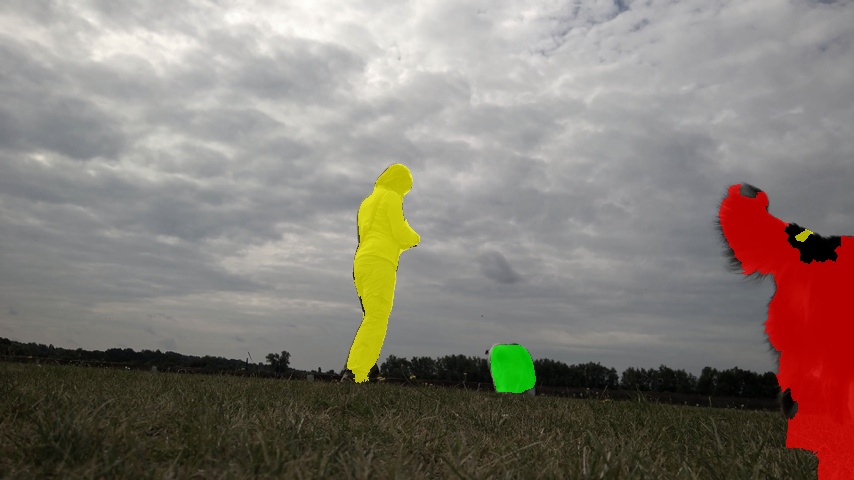} & 
\includegraphics[width = \figqualwidthratioB\linewidth]{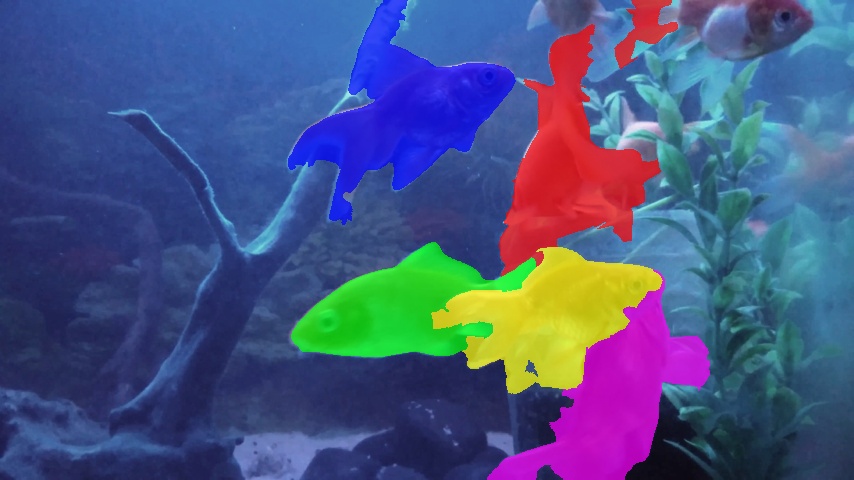} & 
\includegraphics[width = \figqualwidthratioB\linewidth]{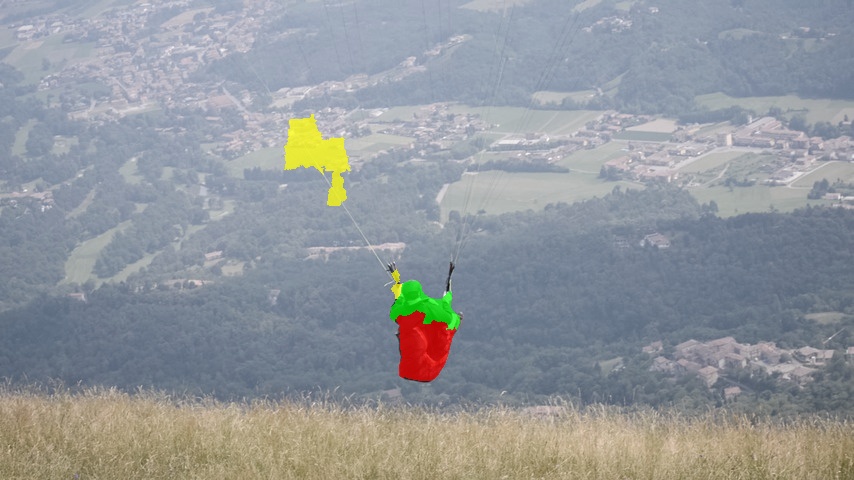} & 
\includegraphics[width = \figqualwidthratioB\linewidth]{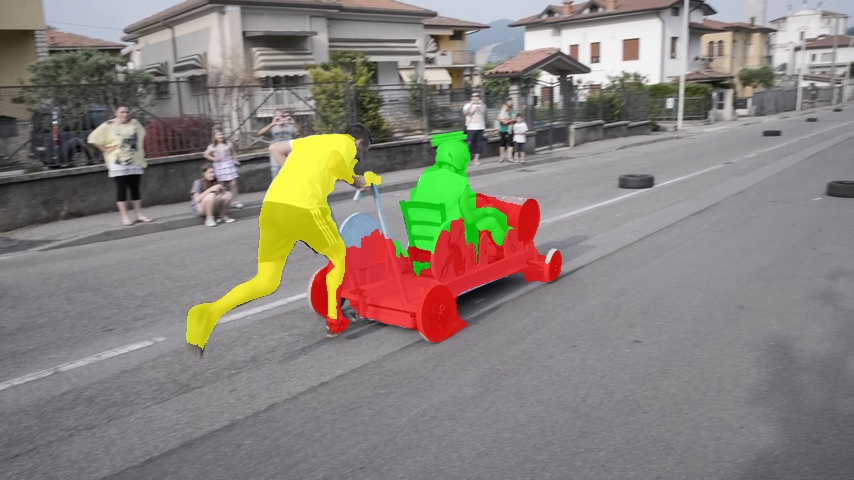}\\

\end{tabular}
\caption{Qualitative results after 8 interaction steps taken from videos of the DAVIS 2017 validation set.}
\label{fig:qual2}
\end{figure*}

To the best of our knowledge, our work is the first to apply GNNs to solve label propagation in interactive VOS. While GNN implementations are less efficient than grid based convolutional layers due to the irregularities of the graph structure, advantages of reduced dimensionality overcome this deficiency and provide a speedy solution for label propagation. Our graph-based solution requires the extraction of node and edge level features which takes up additional computation time. However, most of the feature extraction is a preprocessing step that takes place before the interactive phase and can be easily parallelized, thus it adds a minimal contribution to the waiting time of the annotator.

\section{Method}

The outline of our method is depicted in Fig.~\ref{fig:overview}. The procedure is consistent with the interactive protocol of the DAVIS interactive benchmark. Two steps are repeated in an alternating manner. First, the (simulated) user returns a set of annotation scribbles in one of the frames. Then, our method estimates a full, pixelwise labeling of the video and submits it to the benchmark. After that the two steps are repeated until the end of the session. Starting from the second step of the process, the user provides correctional scribbles in reaction to our estimations in the frame where the error rate of our estimation was the worst. The benchmark measures the time taken by our method to generate the label estimation. The benchmark metrics are computed as a function of accuracy and elapsed time.

Our method relies on a graph constructed over the superpixel segmentation of the video frames. We utilize a GNN to estimate the labeling of the superpixel nodes. In what follows, we describe our approach in detail. First, we provide details about the superpixel segmentation algorithm and the construction of the graph over the superpixel segmentation. Following that, we define our GNN based label propagation model and list the node and edge features used as the GNN input. Finally, we describe the training and inference procedure.

\subsection{Superpixel graph}

One of our main priorities is to construct a quick and efficient method for interactive VOS. For this reason, we stick with a simple and fast superpixel algorithm, the well-known SLIC \cite{superpixel_slic},  which can be easily applied recursively. However, the basic algorithm does not perform that well in terms of boundary recall and undersegmentation error \cite{superpixel_survey}. Nonetheless, objects to be segmented typically move compared to their background or to other objects and optical flow estimation of the direction and magnitude of object movement can be invoked. We use the \textit{FlowNet 2} \cite{flownet2} model. We apply the Canny edge detection algorithm \cite{canny} to find object border candidates. We split superpixel segments with the object border candidates and further apply recursive refinement of the segmentation in the vicinity of these border candidates. This way, the undersegmentation error of the segmentation is significantly reduced. We limit the number of superpixel segments in each frame to have the whole graph fit into GPU memory. Approximate metrics of the graphs can be found in Sect.~\ref{sec:results}.

We construct a graph from the superpixel segmentation, where the segments are interpreted as nodes of the graph. Spatially adjacent or causally connected nodes are linked with edges. Causal connections are, again, extracted from optical flow estimations.

\subsection{Graph Neural Network architecture}

We model the task of video segmentation as a node classification task over the superpixel graph. We learn to estimate label probabilities for each node with a graph neural network (GNN). Our method extract node and edge level features and use them as inputs to our network. We utilize a variant of the Gated GCN \cite{gnn_gated_gcn} architecture, motivated by both of its performance and its ability to maintain an edge representation beyond the ordinary node representations.
We denote the input node feature vectors for node $i$ with $x_i$ and the input edge feature vectors for the directed edge $i \to j$ with $z_{ij}$. More details about the  individual input features follow later (in Sect.~\ref{subsection:method_features}).

First, we use linear embeddings to transform the input node and edge feature vectors to the node and edge representations of the GNN.

$$ h_i^1 = W_n x_i + b_n,\qquad e_{ij}^1 = W_e z_{ij} + b_e $$

\noindent Here, $h_i^l$ denotes the node representation of node $i$ and $e_{ij}^l$ designates the edge representation of edge $i \to j$ in layer $l$. $W_{n} \in \mathbb{R}^{H \times N}, W_e \in \mathbb{R}^{H \times E}, b_n \in \mathbb{R}^H, b_e \in \mathbb{R}^H $ where $N = |x_i|$, $E = |z_{ij}|$ and $H = |h_i|$.

Next, we define the GNN layer based on \cite{gnn_gated_gcn}. The layer relies on a LSTM-like gating mechanism to construct the updated node representation from a weighted mixture of the representation of the adjacent nodes. Individual weights are estimated for each feature.

$$\tilde{e}_{ij}^l = \sigma(C^l e_{ij}^l + D^l h_i^l + F^l h_j^l) $$

$$\tilde{h}_i^{l} = A^l h_i^l + \sum_j{\frac{\tilde{e}_{ij}^l}{\sum_{j^\prime} \tilde{e}_{ij^\prime}^l + \varepsilon} \odot B^l h^l_{j^\prime}}, $$


\noindent where $A^l, B^l, C^l, D^l$ and $F^l$ are all square matrices with a size of $H \times H$. In each layer, a different set of parameters are learned.
We apply batch normalization \cite{batch_norm} and ReLU activations on both the node and edge representations. The input of the layer is added through residual connections \cite{residual} and is followed by the dropout operation:

$$ h_i^{l+1} = \textrm{Dropout}(\textrm{BatchNorm}(\textrm{ReLU}(\tilde{h}_i^{l})) + h_i^l) $$
$$ e_{ij}^{l+1} = \textrm{Dropout}(\textrm{BatchNorm}(\textrm{ReLU}(\tilde{e}_{ij}^{l})) + e_{ij}^l) $$

We stack $L$ GNN layers on top of each other. Finally, we learn a linear classification from the node representations of the last layer to acquire $\{\hat{y}_i\}$, the binary label probabilities for each node:

$$ \hat{y}_i = \sigma(w_y h_i^L + b_y) $$

\subsection{Seed propagation and global label model}

Each layer of the GNN is responsible for the propagation of information from each node to all of the adjacent nodes. In view of this, the range of information propagation within a single inference phase equals the number of layers in the network. In case of static or slow moving objects in images, this range might not be sufficient to connect distant frame pairs of relatively similar contents. One can tackle this problem by  introducing long distance edges into the graph. We applied a simple,  optical flow based seed propagation algorithm and found that it was sufficient for propagating scribble annotations to distant frames of the video. 

As already noted, annotations are given by the user in the form of scribbles. To facilitate the handling of the annotations, we randomly sample seed points from the scribbles. Our seed propagation algorithm propagates seed points to consecutive frames in both directions, as long as optical flow estimations at the location of the seed point are consistent with optical flow estimations from the reversed video. Specifically, the algorithm propagates a seed point $p$ to the consecutive frame if $||g(f(p)) - p||_2 < \beta$, where $f, g$ are the forward and backward optical flow transformations and thus $f(p)$ is the propagated seed point. The backward optical flow is estimated from the consecutive frames of the reversed video. The algorithm tries to propagate each point to both the subsequent and the preceding frames. In the latter case, $f$ and $g$ are swapped. We chose $\beta$ to be $5$ pixels and  found that this value results in a very low false positive rate. While the recall of this method is not very high, the precision is almost perfect: when the propagation of a seed point is allowed to a consecutive frame, results are almost always correct.

Although we successfully extended the range of user annotation propagation, it is not possible to connect intermittently occluded objects before and after the point of occlusion (e.g., a person walking behind trees). To solve this, we utilize a global label model to implicitly store image feature distributions for each label category. We implement the global label model by means of a logistic regression model, which is trained from extracted image feature vectors of randomly selected annotated superpixel segments. Feature vectors are extracted from feature maps estimated by a pretrained \textit{MobileNet v2} \cite{mobilenet} model. The logistic regression model handles multiclass labels using the One-versus-Rest strategy.

\subsection{Node and edge features} \label{subsection:method_features}

We encode information in the node and edge features about the video itself and the present state of the annotation process. The individual features included into the  node representations $x_i$ are as follows:

\begin{itemize}
    \item $x_{i,0},\,  x_{i,1}$ encode the occlusion computed from the forward and backward optical flow estimations. A pixel is considered that it got occluded when no optical flow vectors point to it from the adjacent frame in the reverse direction.
    \item $x_{i,2}$ to $x_{i,7}$ encode the mean and variance of the color distributions regarding the pixels of node $i$ in the CIELAB color space.
    \item $x_{i,8}$ and $x_{i,9}$ store whether user annotations concerning any foreground or background have been given such that they intersect superpixel $i$.
    \item $x_{i,10}$ and $x_{i,11}$ store whether any seeds were propagated into superpixel $i$ with foreground/background labels.
    \item $x_{i,12}$ encodes the foreground/background probability prediction of the global label model.
    \item $x_{i,13}$ encodes the foreground/background probability prediction of the GNN in the previous interaction round.
\end{itemize}

The $z_{ij}$ edge representations consist of the following features:

\begin{itemize}
    \item $z_{ij,0}$ to $z_{ij,2}$ hold the mean $L_2$ color distance of superpixels $i$ and $j$, computed in the CIELAB color space.
    \item $z_{ij,3}$ to $z_{ij,6}$ hold the $L_2$ distance of the mean optical flow vectors of superpixels $i$ and $j$. Both forward and backward optical flow directions are considered.
    \item $z_{ij,7}$ and $z_{ij,8}$ encode whether $i \to j$ is a spatial edge or a causal edge.
\end{itemize}

\subsection{Training procedure} \label{subsection:method_training}

In contrast to semantic segmentation, in VOS, the model is trained to be able to predict boundaries of objects from previously unseen categories, ignoring the semantics. As we must be able to handle previously unseen label configurations with an indefinite number of categories, we choose to approximate the multiclass label estimation task with the combination of several binary label estimation tasks, following the One-versus-Rest strategy. This way, the GNN can be trained to output a single foreground/background probability in each node instead of a probability vector varying in length. Therefore, during prediction of a multiclass label of $k$ categories, $k$ binary prediction rounds must be executed, where each category is selected exactly once as the foreground, while the rest is assigned to be background. The estimated individual binary probabilities are summed and normalized to acquire the final multiclass probability estimations. During training, we randomly partition the label set of the current training sample into two disjoint non-empty subsets. The scribble annotations and the target labels are transformed accordingly.

We train our GNN model against the simulated user provided by the DAVIS interactive benchmark. We start and maintain the state of multiple interactive sessions simultaneously and randomly sample from them to avoid high correlations between consecutive training examples taken from a single session. When preparing a training example, we generate a correctional scribble with the help of the simulated user in response to the previously predicted labels. We sample seed points from the scribbles and perform seed propagation as already described in the beginning of this section and depicted in Fig.~\ref{fig:overview}. In the first interaction step, only foreground objects are annotated by the simulated user. Consequently, we sample background seed points from pixels farther away from foreground annotations. We input all node and edge level features as listed in the previous subsection to the network and infer the binary target label probabilities. We train the network with a binary cross-entropy loss averaged over the nodes. To move the current interactive session forward, we must submit a prediction to it. However, when the current label set consists of more than two categories, the original multiclass prediction cannot be restored from a single binary estimation. In this case a full One-versus-Rest prediction round must be executed. We note, that seed points must be sampled randomly from scribbles with high enough variance, as a fully deterministic procedure could result in the generation of the same interactions and predictions over and over that can jeopardize the success of the training.

\begin{figure*}
    \centering 
    \includegraphics[width=\textwidth]{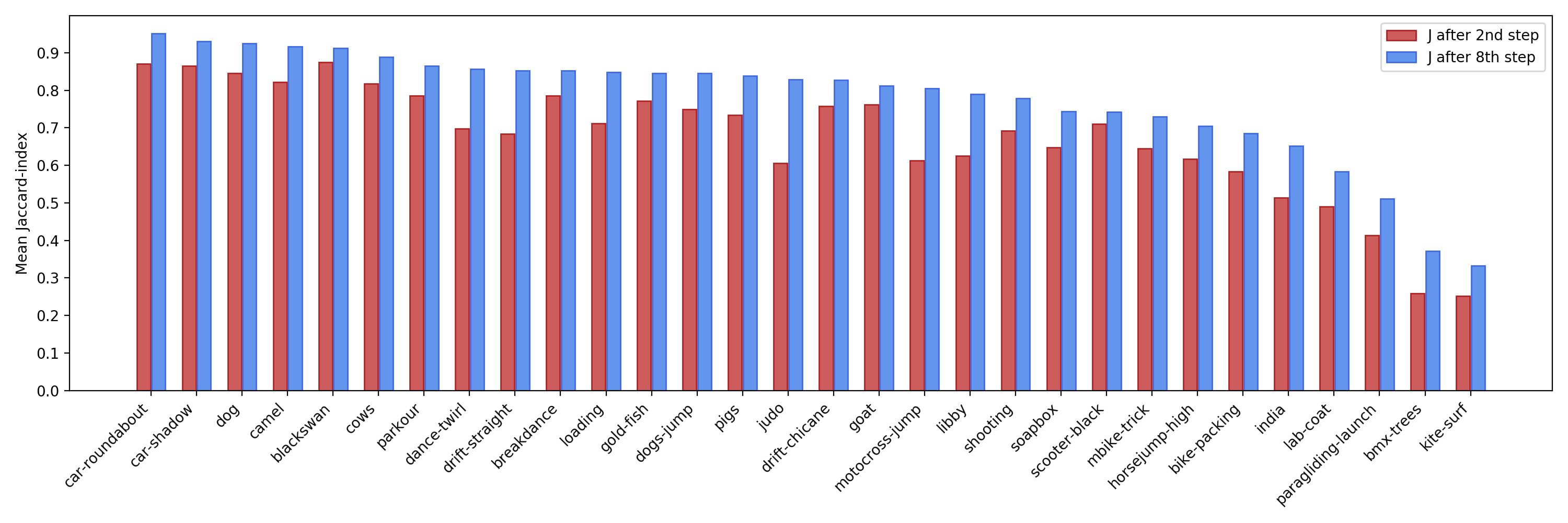}
    \caption{Quantitative results on individual videos of the DAVIS 2017 validation set. Mean Jaccard-scores after the second and eight interaction steps are shown.}
    \label{fig:barchart}
\end{figure*}

\section{Experiments and evaluation} \label{sec:results}

\subsection{Details of the training and prediction processes}

We implement the training process by means of the PyTorch \cite{pytorch} and DGL \cite{gnn_dgl} libraries. We scale the superpixel segmentation to be able to fit the resulting graph into the GPU memory using a $12$ layer GatedGCN network with a hidden representation size of $H = 20$ and a batch size of $1$. We apply dropout to the node representations in each layer with a probability of $0.1$. The mean graph node count is around 60k (about 800 segments per frame), while the mean edge count is around 1.1M directed edges. We use the first $45$ videos of the DAVIS 2017 training set as our training set and the remaining $15$ videos as our validation set. We use the DAVIS 2017 validation set for testing purposes. We note, that the network can be trained by using much fewer videos with a relatively small performance degradation of the final model. We attribute this to the small number of parameters of our network (the $12$ layer network contains a total of 25700 trained parameters). The network can be trained quite quickly; the accuracy saturates after seeing approximately 4000 samples.

We predict the multiclass labels following the One-versus-Rest strategy as already mentioned in Sect.~\ref{subsection:method_training}. During the prediction, one binary inference step is executed for each category in the label set. This approach is significantly faster than that of most contenders of the DAVIS interactive benchmark. On the average, our method takes 45 seconds to finish a full interactive session, consisting of 8 interaction rounds. For the evaluation we use a GTX 1080 Ti graphics card to be able to compare our execution times with other methods. Papers \cite{interactive_oh_cvpr19} and \cite{interactive_ma_net} report similar graphics cards: paper \cite{interactive_oh_cvpr19} completes approximately 5 interactions in a minute, while paper \cite{interactive_ma_net} finishes 7 interactions during this time. It seems that paper \cite{interactive_ma_net} includes preprocessing time into their measurements being taken before revealing the first user interactions for their method. Our preprocessing phase includes the estimation of feature maps, optical flow, and the extraction of node and edge level features from them, which takes approximately 40 seconds for an average video from the dataset. Thus, if we include preprocessing time in the measurement, our method cannot finish 8 interaction rounds within 60 seconds. However, as we have argued, timing measurements should start after the annotator submits their first scribble and a preprocessing phase should be allowed to take place since it does not require the supervision of an annotator at all. At once, our hardware configuration is able to handle approximately $150$ frames. Longer sequences may be simply cut to shorter slices. However, seed propagation can be continued across the boundaries of the slices and the state of the global label model can be maintained indefinitely. These two techniques speed up the annotation of consecutive slices significantly.

\subsection{Interactive VOS results}

We evaluate our approach using the DAVIS 2017 validation set and the simulated user of the interactive benchmark. The benchmark uses (a) the Jaccard-score $\mathcal{J}$, also known as the mean intersection-over-union score and $\mathcal{F}$-score, the measure of contour accuracy \cite{davis_challenge17} in order to compute $\mathcal{J\&F}$ scores, the arithmetic mean of the two averaged over all objects in all frames of the dataset.

Since $\mathcal{J}$ and $\mathcal{J\&F}$ are measured for each object individually and then the object scores are averaged across the whole dataset, videos with many objects are highly responsible for the overall outcome. We summarize the results in Table~\ref{table:main_eval}. We use the standard metrics of the benchmark, AUC $\mathcal{J}$, and AUC $\mathcal{J\&F}$, short for area-under-curve for the $\mathcal{J}$ and $\mathcal{J\&F}$ curves taken over elapsed time, respectively. We split the table into two parts: we list methods with and without extra training data separately. Oh et al. \cite{interactive_oh_cvpr19} measure a 44\% increase in error rate when extra training data is omitted, indicating that methods with many parameters are data-hungry and underperform without the help of additional data.

\begin{table}[htbp]
\caption{Comparison of our method with state-of-the-art interactive methods on the DAVIS 2017 validation set. The rightmost column indicates whether extra training data were used.}
\begin{center}
\begin{tabular}{|c|c|c|c|}
\hline
\textbf{Method} & \textbf{AUC $\mathcal{J}$} & \textbf{AUC $\mathcal{J\&F}$} & \textbf{Extra data} \\
\hline
Heo et al. \cite{interactive20_1st} & \textbf{0.771} & 0.809 & \checkmark \\
\hline
Heo et al. \cite{interactive19_2nd} & 0.704 & - & \checkmark \\
\hline
Oh et al. \cite{interactive_oh_cvpr19} & 0.691 & - & \checkmark \\
\hline
\hline
Miao et al. \cite{interactive_ma_net} & \textbf{0.749} & - & - \\
\hline
Najafi et al. \cite{interactive_najafi_cvpr18} & 0.702 & - & - \\
\hline
Oh et al. \cite{interactive_oh_cvpr19} & 0.555 & - & - \\
\hline
Ours & 0.735 & 0.764 & - \\
\hline
\end{tabular}
\end{center}
\label{table:main_eval}
\end{table}

Our method is superior to most of the contemporary literature, both with or without extra data, except for Heo et al., who use several image and video segmentation datasets for training and surpasses our solution by a margin of 3.5 percents considering the AUC $\mathcal{J}$ metric. Miao et al., is the only one surpassing our solution without extra training data by 1.4 percents. Both papers surpassing ours were published in the past 12 months.

In order to gain some insight into the data structure, we show the error rate of prediction in case of individual videos in Fig.~\ref{fig:barchart} after the second and the eighth interaction steps. It is clear from the figure, that the majority of the videos can be annotated to reach high label accuracy with less than half a minute of total waiting and annotation time from the user. Failures include the \textit{kite-surf} and the \textit{paragliding-launch} sequences, where the segment based solution is unable to correctly segment the cords of the parachutes. We note that most state-of-the-art approaches are also incapable of handling similarly shaped segments. The \textit{lab-coat} and the \textit{bmx-trees} sequences contain object categories where the corresponding segments are relatively small.

To the best of our knowledge, no other graph-based, high performance solutions have been published in interactive VOS. However, until a few years ago, many of the published interactive VOS approaches exploited graph-structures and graphical models, e.g., Markov Random Field (MRF), Bayesian Networks, as cited in Sect.~\ref{subsection:related_ivos}. While the cited publications all use different annotation protocols, we try to compare the performance of our solution to their work by replacing only the GNN of our method to a MRF based node classification algorithm. 

We implement this task by using of the \textit{gco} library \cite{mrf1, mrf2, mrf3}, which optimizes an energy function by means of graph cuts. We define the energy of our graph labeling prediction in the following way:

$$E(\mathcal{L}) = \sum_i U_i(l_i) + \lambda \sum_{i,j} V_{i,j}(l_i, l_j),$$

\noindent where $U_i$ is the unary potential for node $i$ and $V_{i,j}$ is the pair-wise potential for the undirected edge $i \longleftrightarrow j$. $\mathcal{L}$ is a complete labeling of the graph nodes and $l_i$ is the label assigned to node $i$ within the MRF model. We derive $U_i$ from $p_{i,l_i}$, the probability predicted by the global label model for the specific node to be assigned with label $l_i$:

$$U_i(l_i) = -\log p_{i,l_i}$$

Our pair-wise smoothing term is formulated as:

\begin{equation}
  V_{i,j}(l_i, l_j)=\begin{cases}
    0 & l_i = l_j\\
    w_{i,j} & \textrm{otherwise}
  \end{cases}
\end{equation}

Here, $w_{i,j}$ are computed from the strength of the causal connection between nodes $i$ and $j$, which is estimated by optical flow. In other words, when different labels are assigned to nodes with strong causal relation, the energy of the system increases significantly. The weight of the pair-wise term in the energy function was optimized as a hyperparameter.

One advantage of the MRF framework is the capability to handle unseen label sets naturally without the need of the One-versus-Rest strategy. We show the results of this comparison in Table~\ref{table:mrf}. We rely on the same benchmark, dataset and protocol as in the previous case of Table~\ref{table:main_eval}. We measure the average quality of the predicted labeling after the 2nd interaction round and at the end of the interactive sessions, i.e., after the 8th interaction round. The superiority of the neural network approach over classic approaches is convincing.

\begin{table}[htbp]
\caption{Comparison of our method with the Markov Random Field baseline.}
\begin{center}
\begin{tabular}{|c|c|c|}
\hline
\textbf{Method} & \textbf{$\mathcal{J}$ @ 2nd step} & \textbf{$\mathcal{J}$ @ 8th step} \\
\hline
Ours - GNN & 0.622 & 0.741 \\
\hline
Ours - MRF & 0.295 & 0.419 \\
\hline
\end{tabular}
\end{center}
\label{table:mrf}
\end{table}

Next, we analyze our method qualitatively. We show the progression of two interactive labeling sessions in Fig.~\ref{fig:qual1}. Both displayed sessions were chosen from the worst quartile considering the individual error rates of the sequences. In both cases, one or multiple objects appear to be very small in the frame. Although, in some frames the predicted object boundaries are rough, even very small objects happen to be tracked adequately.

The prediction quality after the 8th interaction round in case of five other sequences are shown in Fig.~\ref{fig:qual2}.

\subsection{Ablation study}

We conduct an ablation study to analyze how the removal of key components in our method affect the results. We train and evaluate our approach after disabling the seed propagation algorithm and the global label model individually, then finally both at the same time. We measure the $\mathcal{J}$ metric of the label predictions after 2 and 8 interaction rounds. The results can be seen in Table~\ref{table:ablation1}. Following only two interaction steps, relatively long, unannotated sections still exist in the videos. Under these circumstances, the GNN, limited in range by the number of its layers, simply cannot propagate annotations to major parts of the video on its own. The addition of the seed propagation algorithm results in significant improvement in the case of sequences that are mostly static or contain only slow moving objects. Nonetheless, seed propagation tends to fail in more dynamic scenes.

\begin{table}[h]
\caption{Ablation study to analyze the effect of the removal of certain components of our method. SP and GLM are abbreviations for Seed Propagation and Global Label Model, respectively.}
\begin{center}
\begin{tabular}{|cc|c|c|}
\hline
\textbf{SP} & \textbf{GLM} & \textbf{$\mathcal{J}$ @ 2nd step} & \textbf{$\mathcal{J}$ @ 8th step} \\
\hline
\checkmark & \checkmark & \textbf{0.622} & \textbf{0.741} \\
\hline
- & \checkmark & 0.570 & 0.703 \\
\checkmark & - & 0.485 & 0.712 \\
- & - & 0.209 & 0.639 \\
\hline
\end{tabular}
\end{center}
\label{table:ablation1}
\end{table}

The application of a global label model ensures that even after a single piece of annotation was given by the user, an estimation of the label probabilities are available for each superpixel in the video. However, in several sequences (e.g., \textit{gold-fish}, \textit{pigs}, \textit{lab-coat}), multiple instances of the same semantic categories need to be tracked and segmented. The global label model on its own cannot solve ambiguities between different objects of similar appearance, thus the method fails to tell them apart in frames or regions beyond the information propagation range of the GNN. In conclusion, the seed propagation and the global label model appear to complement each other and the GNN model.

\begin{table}[h]
\caption{Results of our method trained with different subsets of the DAVIS 2017 training set.}
\begin{center}
\begin{tabular}{|c|c|c|}
\hline
\textbf{Training data} & \textbf{$\mathcal{J}$ @ 2nd step} & \textbf{$\mathcal{J}$ @ 8th step} \\
\hline
Original (first 45 videos) & \textbf{0.622} & \textbf{0.741} \\
\hline
First 15 videos & 0.615 & 0.728 \\
First 5 videos & 0.604 & 0.702 \\
\hline
\end{tabular}
\end{center}
\label{table:ablation2}
\end{table}

Despite the severe deterioration experienced after the 2nd interaction round with the removal of components, at the end of the session, the reduction in performance is much more subtle. We attribute this to the fact, that after 8 interaction rounds, the direct propagation of annotation information by the GNN reaches most parts of the video and in this case, the seed propagation and the global label model cease to be the exclusive sources of label information.

We also analyze the robustness of our method to the scarcity of training data. We evaluate our approach trained on only the first $15$ and the first $5$ sequences (taking sequence names in alphabetical order) of the DAVIS 2017 training set. The results are shown in Table~\ref{table:ablation2}. We observe, that our model reaches a relatively high $\mathcal{J}$ score of $0.702$ at the end of the session, even when only $5$ videos, a total of $405$ frames are used for training.

\section{Conclusion and future work}

In this paper, we have shown that a graph neural network based approach is able to achieve state-of-the-art results in interactive VOS with a significantly smaller number of trained parameters and less training data than most existing solutions. Our method utilizes GNNs trained for binary classification with the One-versus-Rest strategy in order to implement multiclass classification of superpixels. We further improve the accuracy by utilizing seed propagation to overcome the limited propagation range of the GNN and a global label model to be able to associate isolated appearances of temporarily occluded objects as well.

A known limitation of our method is the quality of the superpixel segmentation. We implement the splitting of superpixel segments when annotations with multiple different labels intersect a specific segment. We use the Watershed algorithm \cite{watershed} for this purpose. While the Watershed algorithm is far behind the state-of-the-art in image segmentation, in case of splitting small segments, its performance is comparable to contemporary methods and execution is fast. In principle, using Watershed-based segment splitting, our method can reach high segmentation quality, nonetheless this component of our solution could be replaced by other methods. For example, one may consider the refinement of superpixel segmentation based on the GNN label predictions after each interaction round. In spite of the fact that this addition may slightly increase the waiting time of the user, it offers high quality labeling and the decrease of the number of interaction steps.

\bibliographystyle{./bibliography/IEEEtran}
\bibliography{./bibliography/IEEEabrv,./bibliography/refs}

\end{document}